\PassOptionsToPackage{prologue,dvipsnames}{xcolor}
\documentclass[sigconf]{acmart}

\usepackage{amssymb}
\usepackage{amsmath} 
\usepackage{booktabs} 
\usepackage{array}
\usepackage{graphicx} 
\usepackage{multirow}
\usepackage{color}
\usepackage{makecell} 
\usepackage{float}
\usepackage{subcaption}
\usepackage{colortbl}
\usepackage{placeins}
\usepackage{comment}
\usepackage{pifont}
\usepackage{enumitem}
\usepackage{tabularx}
\usepackage{listings}   
\usepackage[utf8]{inputenc}  
\usepackage{pdfpages}
\usepackage{mathtools}      
\usepackage{stmaryrd}       

\usepackage[linesnumbered,vlined,ruled,noend]{algorithm2e}
\usepackage{amssymb}
\SetKwComment{Comment}{$\triangleright$ }{}     
\DeclareMathOperator{\InstValid}{InstValid}
\DeclareMathOperator{\Clean}{Clean}
\DeclareMathOperator{\AtomValidAll}{AtomValidAll}
\DeclareMathOperator{\Consistent}{Consistent}
\DeclareMathOperator{\Complete}{Complete}
\DeclareMathOperator{\IsMathQ}{IsMathQ}
\DeclareMathOperator{\HasMisleadingCue}{HasMisleadingCue}
\DeclareMathOperator{\HasAnswerLeak}{HasAnswerLeak}
\DeclareMathOperator{\LinguisticError}{LinguisticError}
\DeclareMathOperator{\AtomValid}{AtomValid}
\DeclareMathOperator{\Sat}{Sat}

\AtBeginDocument{%
  }

\copyrightyear{2026}
\acmYear{2026}
\setcopyright{cc}
\setcctype{by}
\acmConference[KDD '26]{Proceedings of the 32nd ACM SIGKDD Conference on Knowledge Discovery and Data Mining V.1}{August 09--13, 2026}{Jeju Island, Republic of Korea}
\acmBooktitle{Proceedings of the 32nd ACM SIGKDD Conference on Knowledge Discovery and Data Mining V.1 (KDD '26), August 09--13, 2026, Jeju Island, Republic of Korea}
\acmPrice{}
\acmDOI{10.1145/3770854.3785700}
\acmISBN{979-8-4007-2258-5/2026/08}

\begin{document}

\title{Let's Verify Math Questions Step by Step}



\author{Chengyu Shen}
\authornote{These authors contributed equally to this work.}
\affiliation{%
  \institution{Peking University}
  \city{Beijing}
  \country{China}}
\email{scuuy05@gmail.com}

\author{Zhen Hao Wong}
\authornotemark[1]
\affiliation{%
  \institution{Peking University}
  \city{Beijing}
  \country{China}}
\email{zhenhao1141@stu.pku.edu.cn}

\author{Runming He}
\authornotemark[1]
\affiliation{%
  \institution{Peking University}
  \city{Beijing}
  \country{China}}
\email{hrm1165444624@stu.pku.edu.cn}

\author{Hao Liang}
\authornotemark[1]
\authornote{Project leader.}
\affiliation{%
  \institution{Peking University}
  \city{Beijing}
  \country{China}}
\email{hao.liang@stu.pku.edu.cn}

\author{Meiyi Qiang}
\affiliation{%
  \institution{Peking University}
  \city{Beijing}
  \country{China}}
\email{qiangmeiyi@gmail.com}

\author{Zimo Meng}
\affiliation{%
  \institution{Peking University}
  \city{Beijing}
  \country{China}}
\email{molyheci@stu.pku.edu.cn}

\author{Zhengyang Zhao}
\affiliation{%
  \institution{Peking University}
  \city{Beijing}
  \country{China}}
\email{zhengyangzhao25@stu.pku.edu.cn}

\author{Bohan Zeng}
\affiliation{%
  \institution{Peking University}
  \city{Beijing}
  \country{China}}
\email{13811753844@126.com}

\author{Zhengzhou Zhu}
\authornote{Corresponding author.}
\affiliation{%
  \institution{Peking University}
  \city{Beijing}
  \country{China}}
\email{zhuzz@ss.pku.edu.cn}

\author{Bin Cui}
\affiliation{%
  \institution{Peking University}
  \city{Beijing}
  \country{China}}
\email{bin.cui@pku.edu.cn}

\author{Wentao Zhang}
\affiliation{%
  \institution{Peking University}
  \city{Beijing}
  \country{China}}
\email{wentao.zhang@pku.edu.cn}

\renewcommand{\shortauthors}{Chengyu Shen et al.}

\begin{abstract}
Large Language Models (LLMs) have recently achieved remarkable progress in mathematical reasoning. To enable such capabilities, many existing works distill strong reasoning models into long chains of thought or design algorithms to construct high-quality math question-answer (QA) data for training. However, these efforts primarily focus on generating correct reasoning paths and answers, while largely overlooking the correctness of the questions themselves. 
In this work, we present \textbf{ValiMath}, a benchmark consisting of 2147 human-verified mathematical questions covering a wide range of domains such as arithmetic, algebra, and geometry, which are synthesized and curated from the NuminaMath dataset. Each question is annotated with its logical structure, domain coverage, and question correctness, enabling fine-grained evaluation of question quality. ValiMath serves as a high-quality gold-standard test set for validating mathematical questions in LLM training corpora.
Building upon this benchmark, we further propose \textbf{MathQ-Verify}, a pipeline that performs fine-grained parsing of mathematical questions into atomic assumptions and conclusions, and evaluates their semantic soundness through consistency checks. This pipeline achieves high precision in detecting flawed questions and provides a reliable foundation for cleaning noisy mathematical datasets. Experiments show that MathQ-Verify achieves state-of-the-art performance across multiple benchmarks, improving the F1 score by up to 25 percentage points over the direct verification baseline. MathQ-Verify offers a scalable and accurate solution for curating reliable mathematical datasets, reducing label noise and avoiding unnecessary computation on invalid questions. Our code and data are available at the repository \url{https://github.com/OpenDCAI/MathQ-Verify}.

\end{abstract}

\ccsdesc[500]{Computing methodologies~Machine learning}
\ccsdesc[300]{Computing methodologies~Knowledge representation and reasoning}


\keywords{Data Cleaning, Math Reasoning, Question Verification}


\maketitle


\section{Introduction}
Large Language Models have demonstrated impressive performance across a wide range of tasks~\citep{kaddour2023challengesapplicationslargelanguage, fan2023largelanguagemodelssoftware, openai2024gpt4technicalreport}. In particular, when equipped with chain-of-thought (CoT) of varying lengths~\citep{wei2023chainofthoughtpromptingelicitsreasoning}, some models exhibit the ability to solve moderately complex mathematical problems, such as those from the AIME competition~\citep{aime2025}. This reasoning capability is largely attributed to high-quality data sources and efficient training frameworks~\citep{deepseekai2025deepseekr1incentivizingreasoningcapability}.

However, most existing large scale mathematical question and answer datasets are composed primarily of synthetically generated QA pairs. A critical issue in such datasets is that if the question itself is flawed, such as being ill-posed or logically inconsistent, then the answer cannot be correct—underscoring the fundamental importance of question correctness.

While some recent efforts have focused on verifying the correctness of answers~\cite{song2025prmbench, zheng2024processbench, sun2025mm}, they often neglect to validate the questions themselves, implicitly assuming that all provided problems are well-formed and mathematically sound. This assumption frequently breaks down in real-world or synthetic data scenarios, where generated problems may include internal contradictions or violate foundational mathematical principles~\citep{zhou2024jiuzhang30efficientlyimprovingmathematical}. In such cases, the correctness of the question is uncertain, posing a significant challenge for downstream reasoning tasks. They face the following challenges:

\vspace{0.3em}
\textit{(1) Lack of comprehensive question validation methods.} While several recent studies have begun to examine the correctness of mathematical questions~\citep{liang2025mathcleanbenchmarksyntheticmathematical, li2025questbenchllmsaskright}, their focus is typically limited to a narrow set of error types—such as missing assumptions or vague premises. These efforts fall short of establishing a systematic and comprehensive framework for identifying ill-posed or flawed problems. As a result, many QA datasets, particularly those generated synthetically, still include questions that suffer from internal inconsistencies, logical contradictions, or violations of fundamental mathematical principles.

\vspace{0.3em}
\textit{(2) Lack of stepwise and high-difficulty benchmarks for question verification.} Existing benchmarks, such as MathClean~\citep{liang2025mathcleanbenchmarksyntheticmathematical}, do not provide sufficiently challenging questions, nor do they include fine-grained, stepwise annotations needed to evaluate each stage of a multi-step question verification pipeline. This limits the ability to rigorously assess models' capabilities in detecting and reasoning about complex flaws in mathematical problem formulations.

To advance the rigorous evaluation of mathematical question correctness, we present \textbf{ValiMath}, a stepwise annotated benchmark dataset specifically designed to support fine-grained analysis of question formulation quality. ValiMath extends the existing MathClean benchmark by introducing a broader range of synthetically generated yet realistically flawed math problems, filtered and curated from the NuminaMath dataset~\citep{numina_math_datasets}. From an initial pool of 10,000 auto-generated questions, we sample 2,147 representative problems and annotate each with a sequence of structured correctness labels. These labels span multiple levels of semantic and logical fidelity, capturing both well-formed and defective cases. By offering explicit annotations at each verification step, ValiMath enables systematic and interpretable evaluation of question generation pipelines, and provides a foundation for benchmarking future efforts in automatic data validation.

Building on ValiMath, we further propose \textbf{MathQ-Verify}, a structured verification framework that operationalizes the evaluation of question correctness through five sequential stages: (1) Contaminated Instruction Detection, (2) Linguistic Error Detection, (3) Atomic Condition Error Detection, (4) Cross-condition Conflict Detection, and (5) Condition Completeness Validation. Each stage targets a specific class of flaws commonly observed in synthetic question datasets. Through comprehensive ablation studies, we demonstrate the necessity and effectiveness of each component, establishing MathQ-Verify as a robust reference pipeline for quality assessment in math question generation tasks. In summary, our contribution are as follows:
\begin{itemize}[leftmargin=1.5em,topsep=0.5em]
    \item We construct a new benchmark dataset, \textbf{ValiMath}, which supports fine-grained evaluation of mathematical question correctness. ValiMath contains 2,147 math problems curated from synthetic sources and annotated with structured, stepwise correctness labels that cover a wide spectrum of realistic formulation errors. It provides a principled foundation for evaluating data validation systems in the mathematical reasoning domain.
    
    \item We design a new verification pipeline, \textbf{MathQ-Verify}, which decomposes the validation process into five interpretable stages aligned with formalized criteria. This modular pipeline significantly improves validation performance, achieving state-of-the-art results on MathClean and yielding a 15\% absolute F1 gain over direct baselines on ValiMath.

    \item We conduct extensive experiments to assess the effectiveness of each component within the MathQ-Verify pipeline. Through ablation studies, we systematically validate the individual contribution of each verification stage to the overall performance. Additionally, we demonstrate that incorporating a majority voting strategy across verification outputs significantly enhances precision, achieving up to 90\%, which highlights the robustness and reliability of our proposed approach.
\end{itemize}

\vspace{-0.8em}
\section{Related Work}

\textbf{LLMs and Math Reasoning.}
Large Language Models (LLMs) have demonstrated remarkable capabilities across a wide range of NLP tasks, often surpassing human expert performance in knowledge-intensive domains~\citep{brodeur2024superhumanperformancelargelanguage, Luo_2024}, and generating creative content in writing and design~\citep{Franceschelli_2024, ma2024exploringcapabilitieslargelanguage}. Beyond general-purpose tasks, recent efforts have increasingly focused on enhancing the reasoning abilities of LLMs, aiming to equip models with deeper, more human-like thinking processes for complex problem solving~\citep{patil2025advancingreasoninglargelanguage, yang2025multillmcollaborativesearchcomplex, ren2025deepseekproverv2advancingformalmathematical}. Mathematics, as a domain that demands rigorous logical reasoning, has become a popular benchmark for evaluating these capabilities~\citep{saxton2019analysing, lewkowycz2022minerva,drori2022neural}. Commercial products like ChatGPT~\citep{openai2024openaio1card}, Claude~\citep{claude35sonnet2024} and Gemini~\citep{geminiteam2024geminifamilyhighlycapable} have shown strong performance on math evaluations, while open-source models such as DeepSeek-R1~\citep{deepseekai2025deepseekr1incentivizingreasoningcapability} and the Qwen3 series~\citep{qwen} have also made competitive advances, highlighting the importance of training objectives like reinforcement learning~\citep{bansal2019holist,fawzi2022alphatensor,zhao2023alphageometry}. These advancements have been driven by the availability of high-quality reasoning datasets, especially in math problem solving~\citep{lu-etal-2024-mathgenie,cobbe2021trainingverifierssolvemath,albalak2025bigmathlargescalehighqualitymath,hendrycks2021measuringmathematicalproblemsolving}. Our work enhances this trend by proposing a new benchmark to evaluate the models' capability of filtering and verifying the potential incorrect math questions.

\noindent\textbf{Synthetic Data.} 
Training LLMs requires massive amounts of data~\citep{yang2024qwen25mathtechnicalreportmathematical, abdin2024phi4technicalreport}, but existing web corpora have become increasingly insufficient to support full training pipelines, particularly in the pretraining stage~\citep{villalobos2024rundatalimitsllm,perełkiewicz2024reviewchallengesmassivewebmined}. In addition, many open-source datasets suffer from noisy supervision and inconsistent quality, making them suboptimal for model training~\citep{jiang2024investigatingdatacontaminationpretraining}. To address this, recent work has explored synthetic data generation using LLMs~\citep{nadas2025syntheticdatagenerationusing,qin2025scalinglawssyntheticdata}, enabling controllable data distributions and customizable supervision signals. However, synthetic data especially in the domain of mathematical question generation often contains subtle or critical errors in both questions and solutions~\citep{lu2024mathgenie,vendrow2025platinum,shen2025mathqverify,wang2023mathshepherd}, highlighting the need for rigorous verification before such data can be reliably used for training or evaluation. Based on this, we explore methods as a reference for verifying and filtering erroneous mathematical data.

\noindent\textbf{Verification in Math QA.} 
Much attention has been given to improving the accuracy of LLMs on math problem answering, including developing better reasoning strategies and addressing common model failure modes~\citep{xu2024chatglmmathimprovingmathproblemsolving, zhang-etal-2024-rationales}. Models fine-tuned with reinforcement learning have even demonstrated the ability to solve competition-level math problems~\citep{deepseekai2025deepseekr1incentivizingreasoningcapability,openai2024openaio1card,bercovich2025llamanemotronefficientreasoningmodels}. However, when presented with flawed or ill-posed questions, LLMs often lack the ability to detect inconsistencies or exhibit overthinking behaviors~\citep{ouyang2025treecutsyntheticunanswerablemath,fan2025missingpremiseexacerbatesoverthinking}. Recent efforts such as MathClean~\citep{liang2025mathcleanbenchmarksyntheticmathematical} and QuestBench~\citep{li2025questbenchllmsaskright} have introduced benchmarks for identifying flawed problems and missing assumptions, revealing substantial gaps in model robustness. Despite these advances, structured and comprehensive verification for math problems has yet to receive broad attention. To bridge this gap, we introduce \textbf{ValiMath}, a benchmark dataset tailored for stepwise verification of question correctness. To showcase its utility, we provide a formalized multi-stage verification pipeline as a reference approach, illustrating how semantic decomposition enables fine-grained quality assessment.

\begin{figure*}
\centering 
\includegraphics[width=
0.98\textwidth]{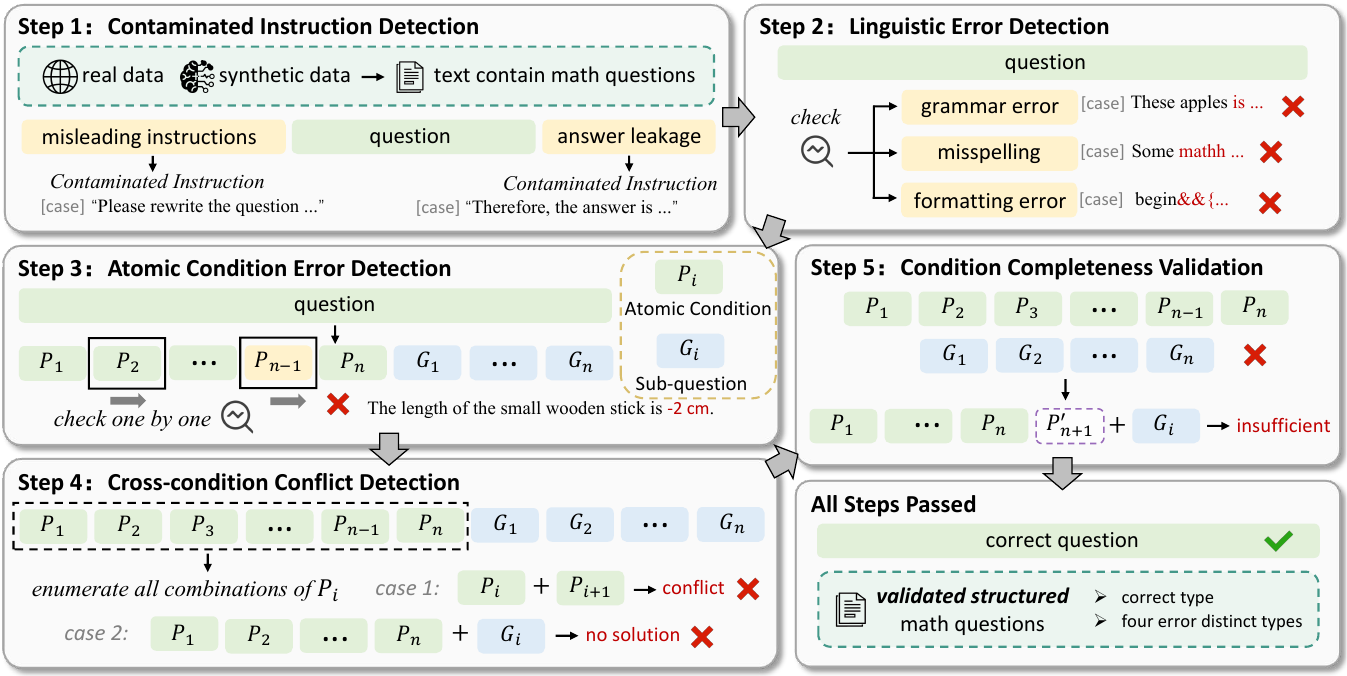} 
\caption{An overview of our MathQ-Verify framework. Given a math question, we extract atomic conditions \(P_i\) and the goal \(G_i\), and conduct a five-step verification process: (1) detection of contaminated instructions, (2) linguistic error detection, (3) atomic condition verification, (4) cross-condition conflict detection, and (5) condition completeness validation. Only questions that pass all checks are retained as correct structured math questions.}
\label{fig:overview}
\end{figure*}

\section{ValiMath: The Question Verification Bench}\label{ValiMath}

\subsection{Dataset Overview}
While MathClean\citep{liang2025mathcleanbenchmarksyntheticmathematical} provides a benchmark for validating math question correctness, it primarily contains simpler questions and lacks fine-grained annotations. To introduce a more diverse range of difficulties for comprehensive evaluation, we present \textbf{ValiMath}, a benchmark of 2,147 questions (1,299 correct, 848 incorrect) covering five diverse error types (defined in Section~\ref{sec:method}) as shown in Table~\ref{tab:error_types}.

\begin{table}[h]
  \centering
  \caption{Data distribution of ValiMath.}
  \vspace{-5pt}
  \label{tab:error_types}
  \resizebox{0.65\linewidth}{!}{%
    \begin{tabular}{lrr}
      \toprule
      \textbf{Category}          & \textbf{Count} & \textbf{\%} \\
      \midrule
      Instruction Error          & 143  & 6.66  \\
      Nonsemantic Error          & 119  & 5.54  \\
      Minimal Domain Error       & 171  & 7.96  \\
      Contradiction Error        & 300  & 13.97 \\
      Completeness Error         & 115  & 5.36  \\
      \midrule
      \textbf{Correct Instances} & 1299 & 60.50 \\
      \bottomrule
    \end{tabular}%
  }
  \vspace{-4pt}
\end{table}

Our dataset encompasses a wide range of difficulty levels and diverse mathematical domain labels. The correctness of the data is ensured through a rigorous filtering and validation process. We have reorganized and developed a new secondary classification system based on the MSC classification standard~\citep{MSC2020}, which has been systematically applied to our data. The dataset is constructed through a meticulous multi-stage pipeline, involving mathematics domain experts to guarantee high-quality annotations. This comprehensive process ensures both the correctness of the questions and the accuracy of the labels. More details and the complete data construction process are provided as follows.

\subsection{The Construction of ValiMath}
\label{VailiMath_construction}
This subsection introduces the construction process of the ValiMath.

\textbf{Step1: Data Synthesis.} 
For synthetic data in the mathematics domain, especially for the generation of questions, there is a high demand for diversity. Additionally, the synthetic data itself tends to align more closely with the distribution of model training data. To construct a diverse set of questions, while also introducing more challenging problems compared to MathClean~\citep{liang2025mathcleanbenchmarksyntheticmathematical}, we thoroughly researched existing prompts used for generating synthetic questions. Based on our findings, we carefully designed a prompt (provided in repository) that ensures the diversity of the generated questions. To ensure the generated mathematical problems possess a certain level of difficulty and challenge, we ultimately used NuminaMath~\citep{numina_math_datasets} as our source data for question synthesis.

Furthermore, to capture a broad range of potential errors that the model might generate, we employed models of different sizes for question synthesis. Stronger reasoning models tend to generate problems with a lower error rate during normal question synthesis, while smaller models are more likely to produce erroneous questions~\citep{wei2022emergent,honovich2022selfinstruct,lu2024mathgenie,chowdhery2022palm}. Therefore, we used a variety of models, including: Qwen2.5-3B-Instruct, Qwen2.5-7B-Instruct, Qwen2.5-72B-Instruct, GPT-4o, and GPT-o3mini, ensuring that each model generated a similar number of questions. In total, we synthesized 10,000 mathematical questions in this stage.

\textbf{Step2: Filtering.}
To ensure the uniqueness and determinacy of the final answers, we applied a GPT-4o-based prompt filtering process to exclude all multiple-choice questions. It is important to note that, since our objective is to construct a dataset containing both correct and incorrect questions, no additional filtering was performed beyond this step. This preserves the natural distribution of model-generated outputs.

\textbf{Step3: Data Selection.}
To reduce annotation costs in advance, we observed during preliminary experiments that general-purpose models such as GPT-4o tend to generate more correct than incorrect mathematical questions in our raw data (with fewer than 30\% incorrect samples). To address this, we first performed a coarse-grained filtering step to retain a higher proportion of incorrect questions for annotation.

Specifically, we employed Qwen2.5-72B-Instruct to directly assess the correctness of each generated question using the baseline judgment method described in the main paper. Based on these results, we curated a subset of approximately 4,000 mathematical questions with a target ratio of correct to incorrect samples set at 1:3. This step effectively reduced downstream annotation workload while preserving as many erroneous questions as possible, thereby enhancing the effectiveness of subsequent data verification.

\textbf{Step4: Expert Annotation.}
After obtaining the initial set of 4,000 mathematical questions, we invited domain experts in mathematics to perform detailed annotations. These annotators were selected through a three-stage screening process from an initial pool of approximately 150 candidates. They were evaluated based on the correctness of their annotations on 10, 50, and 100 questions, respectively. Ultimately, 8 experts were chosen. After a month of intensive training, these experts fully met our standards for annotation quality and domain expertise. In addition, their ethical integrity and moral standards were also assessed and confirmed to meet our predefined requirements. The goal was to ensure the accuracy of the binary correctness labels ("correct" or "incorrect") and, for incorrect questions, to further annotate the specific steps where errors occurred, following the five-stage framework defined in MathQ-Verify. The annotation process proceeded as follows:

\begin{enumerate}[leftmargin=1.5em, topsep=0.5em]
    \item \textit{Annotation Guidelines.} Experts were provided with comprehensive annotation instructions (detailed in Appendix~\ref{appendix:annotation}), including precise definitions of each step in the MathQ-Verify process as outlined in the main paper. Each question was evaluated across the five reasoning stages, and labeled accordingly as either “correct” or “incorrect.” For incorrect questions, annotators further specified the earliest step at which the error occurred.
    \vspace{0.3em}
    \item \textit{Independent Dual Annotation.} Each question was independently annotated by at least two domain experts to mitigate individual biases. Discrepancies between annotations were resolved through a final joint review process, during which experts reached a consensus based on predefined labeling criteria.
    \vspace{0.3em}
    \item \textit{Cross-validation.} Following the dual annotation stage, we conducted a round of cross-validation to verify the reliability of the annotations. This involved a secondary review of previously labeled data, focusing on samples with disagreement, to ensure consistency and correctness in the final labels.
    \vspace{0.3em}
    \item \textit{Subjectivity Analysis.} To further assess and control for annotator subjectivity, we randomly sampled a subset of annotated questions for inter-annotator agreement analysis. We applied standard consistency metrics to quantify labeling stability across experts and identify any systemic annotation biases.
    \vspace{0.3em}
    \item \textit{Quality Assurance.} Through the above multi-step annotation workflow, we ensured the reliability, consistency, and accuracy of the labeled dataset, providing a high-quality foundation for dataset construction and downstream evaluation.
\end{enumerate}

\section{MathQ-Verify} \label{sec:method}

In this section, we introduce our MathQ-Verify, as shown in Figure~\ref{fig:overview}, designed to verify the correctness of mathematical questions. Our methodology systematically deconstructs each question into its fundamental components and evaluates them step by step through a carefully designed pipeline. This pipeline includes multiple stages, including structural analysis, logical consistency checks, and contextual validation, to ensure a comprehensive assessment of the question's quality and reasonableness. By rigorously analyzing and verifying each question, our approach aims to establish a robust framework for evaluating the correctness and soundness of mathematical questions.

\subsection{Problem Formulation}
We define the set of questions in the mathematical QA dataset as $\mathcal{Q} = \{q_i\}_{i=1}^L$, where $L$ is the number of questions and each $q_i$ denotes a natural language math question. \textbf{A valid question}, called correct question as well, means it has clear conditions, is well-defined and free from contradictions, and can be solved, and the invalid question is on the contrary side. This set can be partitioned into two disjoint subsets: a set of well-formed (valid) questions $\mathcal{Q}^{\text{valid}}$ and a set of ill-formed (invalid) questions $\mathcal{Q}^{\text{invalid}}$, such that:
\begin{align*}
	\mathcal{Q} = \mathcal{Q}_{\text{valid}} \cup \mathcal{Q}_{\text{invalid}}, \quad \mathcal{Q}_{\text{valid}} \cap \mathcal{Q}_{\text{invalid}} = \emptyset
\end{align*}

Identifying whether a given question is valid is crucial for downstream tasks such as dataset refinement. Therefore, we aim to learn a verification function $V: \mathcal{Q} \rightarrow \{0,1\}$, where $V(q_i) = 1$ indicates that $q_i$ is a valid question, and $V(q_i) = 0$ otherwise.

To determine whether a question $q_i$ is valid, we propose to decompose it into two structured components using LLM (detailed in repository), which are then used as the basis for verification through $V(q_i)$. Specifically:

\ding{182} \textbf{Atomic Conditions}, $\mathcal{P}(q_i) = \{P_1, P_2, \dots, P_N\}$: These represent the minimal and indivisible conditions in the question, serving as the building blocks for a complete math question.

\ding{183} \textbf{Target Goals}, $\mathcal{G}(q_i) = \{G_1, G_2, \dots, G_M\}$: These define the specific objectives or queries that the question aims to address, encapsulating its intended purpose.

These two components are systematically combined in text form to reconstruct the complete question statement, enabling a rigorous evaluation of its correctness and logical coherence.

\begin{figure}[h]
\centering
\vspace{-1mm}
\includegraphics[width=0.8\linewidth]{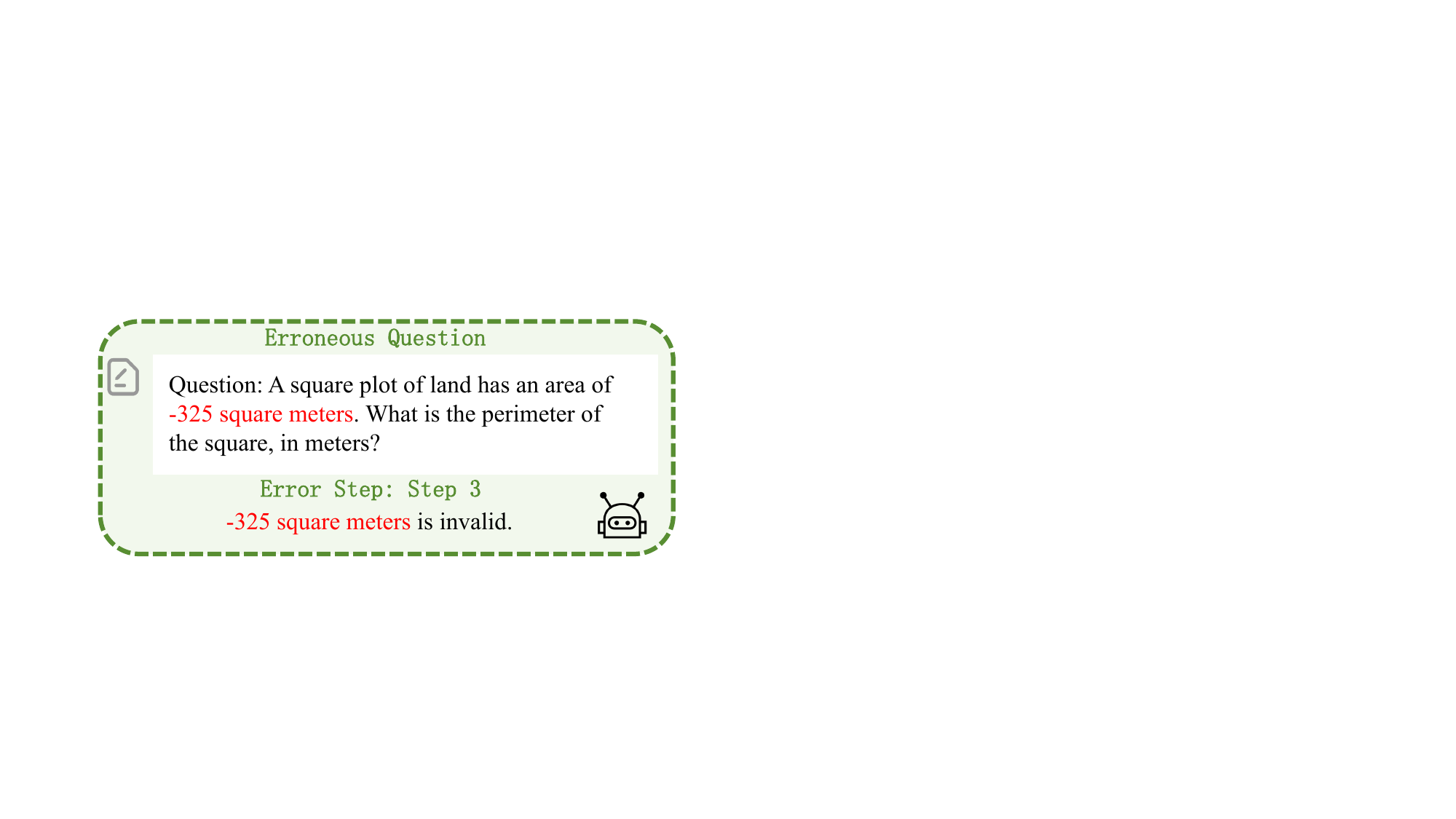}
\caption{An example of an incorrect atomic condition: a negative area contradicts mathematical definitions.}
\label{fig:case}
\vspace{-4mm}
\end{figure}

\subsection{Question Correctness Verification}
\label{subsec:verify}
We verify the correctness of a mathematical question $q$ using a five-stage pipeline, \emph{MathQ-Verify}, which decomposes $q$ into formal components and checks them for instruction integrity, linguistic soundness, atomic validity, global consistency, and goal completeness. This subsection provides (i) precise notation, (ii) five predicates that formalize each stage, and (iii) a single overall decision rule $V(q)\in\{0,1\}$, Formal algorithm is provided in Appendix~\ref{appendix:algo}.


\vspace{0.3em}
\textbf{Step 1: Contaminated Instruction Detection.}
We first ensure that $q$ is a genuine math question and that its phrasing does not bias, leak, or otherwise contaminate the problem-solving objective. Typical violations include meta-instructions (e.g., ``please rewrite'' / ``please rephrase'') and explicit answers embedded in the prompt. Formally, we define $\mathrm{InstValid}(q)=True$ if $q$ does not have any violation instructions above.

\vspace{0.3em}
\textbf{Step 2: Linguistic Error Detection.}
A question $q$ is considered \emph{clean} if it has no semantic imperfections, including spelling mistakes, grammar errors, and Latex anomalies, that hinder readability or parsing while leaving the mathematical content intact. Then we define $\mathrm{Clean}(q) = True$ if there is no semantic imperfections mentioned above.

\vspace{0.3em}
\textbf{Step 3: Atomic Condition Error Detection.}
We decompose a "clean" (defined in Step 2) question into several minimal conditions, called atomic conditions, and do some checks below directly using prompting LLM. The prompt is provided in the repository.
Each atomic condition $P_j\in\mathcal{P}(q)$ must conform to the basic rules of the relevant domain (algebra, geometry, analysis, etc.). We check \emph{atomic validity} at the statement level (e.g., rejecting definitions that contradict standard semantics, such as treating a discrete variable as continuous without qualification). If all the conditions in $q$ obeys all these mathematical rules above, then $\mathrm{AtomValidAll}(q)=True$.

\vspace{0.3em}
\textbf{Step 4: Global Consistency Checking.}
Beyond the validity of individual atoms, the full condition set $\mathcal{P}(q)$ must be jointly satisfiable. Instead of quantifying over all nontrivial subsets (which is cumbersome and unnecessary), we require a single global condition $\mathrm{Sat}(\mathcal{P}(q))$. By prompting LLM (provided in repository), if the model outputs ``no contradictions'' at the level of the entire premise set then we define $\mathrm{Consistent}(q)=True$.

\textbf{Step 5: Condition Completeness Evaluation.}
After obtaining the consistent $\mathcal{P}(q)$, we check the completeness by LLM (provided in repository) that every target in $\mathcal{G}(q)$ should be derivable from the available conditions, then $\mathrm{Complete}(q)=True$. If some $G\in\mathcal{G}(q)$ is not entailed (cannot derive answer from premises), then the question is under-specified (missing information).

Therefore, the overall decision for q is determined as follows:
\begingroup
\setlength{\abovedisplayskip}{4pt}
\setlength{\belowdisplayskip}{4pt}
\setlength{\abovedisplayshortskip}{4pt}
\setlength{\belowdisplayshortskip}{4pt}
\begin{equation}
\label{eq:overall-decision}
\begin{aligned}
V(q)\ & =\ \mathbf{1}\Big[
  \mathrm{InstValid}(q)\ \wedge\ \mathrm{Clean}(q)\ \wedge\ \mathrm{AtomValidAll}(q)\  \\ & \quad \quad \quad \wedge\ \mathrm{Consistent}(q)\ \wedge\ \mathrm{Complete}(q)
\Big]
\end{aligned}
\end{equation}
\endgroup
\noindent\textit{The following are some illustrative examples.}
\begin{enumerate}[leftmargin=1.8em,topsep=0.5em]
  \item \textbf{Contradiction (fails Step 4).} Let $\mathcal{P}(q)=\{x\in\mathbb{Z},\ x\ \text{is even},$ $\ x=2k+1\}$ with $k\in\mathbb{Z}$. The set is not satisfiable, so $\mathrm{Consistent}(q)$ fails, hence $V(q)=0$ by~\eqref{eq:overall-decision}.
  \item \textbf{Under-specification (fails Step 5).} If $\mathcal{P}(q)=\{\text{``$n$ is a} \text{polygon''}\}$ and $\mathcal{G}(q)=\{\text{``compute the sum of interior angles''}\}$, but the $\mathcal{P}(q)$ does not have the numeric value of polygon sides. Thus $\mathrm{Complete}(q)$ fails and $V(q)=0$.
  \item \textbf{Valid (passes all steps).} If $\mathcal{P}(q)=\{x\in\mathbb{Z},\ x\ \text{is even},\ 2x+3=19\}$ and $\mathcal{G}(q)=\{\text{solve for }x\}$, then the atoms are valid and jointly satisfiable; moreover, $x=8$ is derivable from $\mathcal{P}(q)$. Hence $V(q)=1$.
\end{enumerate}

\textbf{Remarks on implementation.}
The five predicates abstract away from any particular detector or solver; concrete implementations (e.g., rule-based checks, symbolic solvers, or verifiers) can be described in the implementation section. Keeping the formal layer model-agnostic improves clarity and reproducibility, while~\eqref{eq:overall-decision} provides a single, auditable criterion for acceptance.

\vspace{-1em}
\subsection{Multi-Expert Voting Strategy}
\definecolor{InvBG1}{HTML}{FEF9F6}
\definecolor{InvBG2}{HTML}{FBEDE4}
\definecolor{InvBG3}{HTML}{F7E1D3}
\definecolor{InvBG4}{HTML}{F3D6C3}
\definecolor{InvBG5}{HTML}{EFCCB2}
\definecolor{greenbg}{HTML}{B2DCC6}

\begin{table*}
\captionsetup{justification=centering}
  \centering
  \caption{MathQ-Verify and baseline performance on MathClean-Simple and MathClean-Challenging. \\ Inv.$^\Diamond$ denotes answers requiring GPT-4o-based semantic parsing; green highlights indicate the best scores, same for later tables.}
  \label{tab:gsm8k_math}
  \vspace{-1em}
  \setlength{\tabcolsep}{5pt}
  \renewcommand{\arraystretch}{1}
  \resizebox{0.9\textwidth}{!}{%
    \begin{tabularx}{\textwidth}{%
        c  
        c  
        *{3}{>{\centering\arraybackslash}X}
        *{3}{>{\centering\arraybackslash}X}
      }
      \toprule
      \multirow{3}{*}{\textbf{Model}}
        & \multirow{3}{*}{\textbf{Method}}
        & \multicolumn{3}{c}{\textbf{MathClean‐Simple}}
        & \multicolumn{3}{c}{\textbf{MathClean‐Challenging}} \\
      \cmidrule(lr){3-5}\cmidrule(lr){6-8}
        & 
        & \textbf{Acc (\%)} 
        & \textbf{F1 (\%)} 
        & \textbf{Inv.}$^\Diamond$ (\#) 
        & \textbf{Acc (\%)} 
        & \textbf{F1 (\%)} 
        & \textbf{Inv.}$^\Diamond$ (\#) \\
      \midrule
      \multirow{2}{*}{Qwen2.5-7B}
        & Baseline & 67.75 & 74.02 &  9 & 71.33 & 76.13 &  20 \\
        & MathQ-Verify      & \textbf{74.38} & \cellcolor{greenbg}{\textbf{76.09}} &  4 & \textbf{76.75} & \cellcolor{greenbg}{\textbf{77.06}} &  11 \\
    \addlinespace[0.4ex]
    \multirow{2}{*}{Qwen2.5-Math-7B}
        & Baseline & 67.63 & 71.94 &  99 & 68.00 & 71.53 &  330 \\
        & MathQ-Verify      & 59.50 & 68.02 &  \textbf{183} & 56.08 & 61.39 &  {\textbf{543}} \\
    \addlinespace[0.4ex]
    \multirow{2}{*}{Llama-3.1-8B}
        & Baseline & 59.62 & 46.96 &  2 & 64.88 & 58.82 &  5 \\
        & MathQ-Verify      & 65.62 & 72.55 &  22 & 65.50 & 72.42 &  4 \\
    \addlinespace[0.4ex]
    \multirow{2}{*}{DeepSeek-R1-Distill-Qwen-7B}
        & Baseline & 62.38 & 59.16 &  37 & 65.25 & 63.61 &  214 \\
        & MathQ-Verify      & 66.88 & 73.76 &  26 & 68.04 & 74.10 &  133 \\
      \bottomrule
    \end{tabularx}%
  }
\end{table*}

To enhance the robustness of condition verification, we propose a \textit{multi-expert voting strategy} that aggregates predictions from multiple independently models. Instead of relying on a single model’s output, this strategy leverages \textit{majority voting} to reach a collective decision. Formally, we consider configurations denoted as \textbf{$(n, k)$}, where \textbf{$n$} is the total number of models in voting, and \textbf{$k$} is the minimum number of models that must agree (i.e., output a positive decision) for the final prediction to be accepted.

Let $V_j(q_i) \in \{0,1\}$ denote the binary verification output of the $j$-th model on query $q_i$. The final decision $V^{\text{final}}(q_i)$ is given by:
\begin{equation}
V^{\text{final}}(q_i) = 
\begin{cases}
1, & \text{if } \sum\limits_{j=1}^{n} V_j(q_i) \geq k \\
0, & \text{otherwise}
\end{cases}
\end{equation}

This formulation provides a tunable mechanism to balance precision and recall by adjusting $k$ relative to $n$. By requiring agreement among a subset of models, the strategy mitigates individual model biases and reduces variance in predictions, thereby increasing the overall reliability of the verification process. The empirical effectiveness of this approach is demonstrated in Section~\ref{sec:experiments}.

\section{Experiments}\label{sec:experiments}
In this section, we conduct a series of rigorous experiments to systematically evaluate the effectiveness and impact of our pipeline. Specifically, we aim to address the following research questions:

\textbf{Q1:} How does MathQ-Verify perform on established benchmarks, and is the pipeline adaptable across different model architectures?

\textbf{Q2:} To what extent does majority voting enhance precision, thereby increasing the proportion of correctly identified questions within the filtered dataset?

\textbf{Q3:} Does the filtering mechanism employed by our pipeline introduce substantial distributional shifts in the resulting dataset?

\subsection{Experimental Setup}

\noindent\textbf{Models.}
We evaluate a total of 14 large language models in this study, encompassing both reasoning-oriented and non-reasoning models. All models are \textbf{Instruct version} if only labeled with parameter size. The reasoning-oriented models, such as DeepSeek-R1-671B, Gemini-2.5-Pro, QwQ-32B and GPT-o4-mini, are explicitly optimized for handling complex tasks that require multi-step inferential reasoning. In contrast, the non-reasoning models, including Qwen2.5-7B, Qwen2.5-Math-7B, Qwen2.5-72B, Qwen2.5-Math-72B, Llama-3.1-8B, S1.1-32B, Llama-3.1-70B, and GPT-4o, are primarily designed for general-purpose language understanding or specialized mathematical reasoning without explicit support for extended reasoning chains.

\noindent\textbf{Dataset.}
We utilize the synthesized and annotated versions of mathematical data from the MathClean benchmark as our primary evaluation datasets. Additionally, we incorporate the ValiMath dataset, as described in Section~\ref{ValiMath}, to comprehensively assess the performance of our approach.

\definecolor{graybg}{HTML}{F5F5F5}
\definecolor{InvBG1}{HTML}{FEF9F6}
\definecolor{InvBG2}{HTML}{FBEDE4}
\definecolor{InvBG3}{HTML}{F7E1D3}
\definecolor{InvBG4}{HTML}{F3D6C3}
\definecolor{InvBG5}{HTML}{EFCCB2}
\definecolor{greenbg}{HTML}{B2DCC6}

\begin{table*}
\captionsetup{justification=centering}
  \centering
  \small
  \caption{Comparison of our MathQ-Verify and baseline performance on ValiMath. \\ S$k$ denotes evaluating the accuracy as the same of $k$-th step to the entire dataset directly, rather than checking step by step.}
  \label{tab:eight_metrics_beautified}
  \vspace{-1em}
  \renewcommand{\arraystretch}{1.0}
  \setlength{\tabcolsep}{4pt}
  \resizebox{0.96\textwidth}{!}{%
    \begin{tabularx}{\textwidth}{
      >{\centering\arraybackslash}p{30mm}
      >{\centering\arraybackslash}p{13mm}
      *{3}{>{\centering\arraybackslash}X}
      >{\centering\arraybackslash}p{10mm}
      *{4}{>{\centering\arraybackslash}X}
    }
      \toprule
      \textbf{Model} & \textbf{Method} & \textbf{Acc} & \textbf{F1} & \textbf{Prec} & \textbf{Inv.$^\Diamond$} & \textbf{S1\&S2} & \textbf{S3} & \textbf{S4} & \textbf{S5} \\
      \midrule
      \multicolumn{10}{c}{\textbf{Non-reasoning Models}} \\
      \midrule
      \multirow{2}{*}{Qwen2.5-7B}
        & Baseline & 67.49 & 76.08 & 68.56 &  1 & 73.64 & 77.32 & 69.54 & 73.22 \\
        & Ours     & 67.91 & 77.61 & 67.15 &  4 & 78.71 & 81.56 & 75.83 & 80.25 \\
      \addlinespace[0.4ex]
      \multirow{2}{*}{Qwen2.5-Math-7B}
        & Baseline & 64.14 & 71.92 & 68.33 &  377 & 63.48 & 67.63 & 68.05 & 66.60 \\
        & Ours     & 60.18 & 67.77 & 66.40 &  \textbf{615} & 60.92 & 62.93 & 63.81 & 61.71 \\
      \addlinespace[0.4ex]
      \multirow{2}{*}{Qwen2.5-72B}
        & Baseline & 70.28 & 78.65 & 69.57 &  1 & 73.92 & 80.39 & 75.41 & 76.57 \\
        & Ours     & 72.38 & 80.77 & 69.79 &  0 & \textbf{81.60} & \textbf{83.98} & 75.55 & \textbf{80.53} \\
      \addlinespace[0.4ex]
      \multirow{2}{*}{Qwen2.5-Math-72B}
        & Baseline & 68.28 & 76.70 & 69.03 &  75 & 68.37 & 77.36 & 74.90 & 74.57 \\
        & Ours     & 68.37 & 77.39 & 68.19 &  44 & 73.40 & 80.25 & \textbf{76.29} & 76.53 \\
      \addlinespace[0.4ex]
      \multirow{2}{*}{Llama-3.1-8B}
        & Baseline & 59.52 & 70.87 & 62.77 &  2 & 72.47 & 75.87 & 71.36 & 75.13 \\
        & Ours     & 63.25 & 73.43 & 65.27 &  41 & 72.66 & 76.06 & 71.91 & 75.97 \\
      \addlinespace[0.4ex]
      \multirow{2}{*}{Llama-3.1-70B\textsuperscript{\ding{177}}}
        & Baseline & 66.84 & 75.26 & 68.59 &  2 & 71.22 & 74.99 & 68.51 & 72.75 \\
        & Ours     & 67.26 & 75.72 & 68.67 &  4 & 72.66 & 76.06 & 71.91 & 75.97 \\
      \addlinespace[0.4ex]
      \multirow{2}{*}{S1.1-32B}
        & Baseline & 67.77 & 72.76 & 74.46 &  62 & 54.63 & 63.25 & 64.04 & 59.25 \\
        & Ours     & 75.31 & 78.78 & 82.07 &  8 & 63.67 & 61.11 & 59.94 & 58.13 \\
      \addlinespace[0.4ex]
      \multirow{2}{*}{GPT-4o\textsuperscript{\ding{176}}}
        & Baseline & 66.60 & 71.74 & 73.51 &  13 & 58.97 & 62.83 & 58.87 & 58.92 \\
        & Ours     & 72.29 & 77.52 & 76.11 &  6 & 70.33 & 66.65 & 60.18 & 63.48 \\
      \midrule
      \multicolumn{10}{c}{\textbf{Reasoning Models}} \\
      \midrule
      \multirow{2}{*}{GPT-o4-mini\textsuperscript{\ding{175}}}
        & Baseline & 73.31 & 77.59 & 78.86 &  0 & 58.22 & 64.42 & 65.21 & 61.25 \\
        & Ours     & \textbf{79.23} & \cellcolor{greenbg}\textbf{83.36} & 80.88 &  4 & 73.82 & 68.84 & 64.32 & 65.21 \\
      \addlinespace[0.4ex]
      \multirow{2}{*}{Gemini-2.5-Pro\textsuperscript{\ding{173}}}
        & Baseline & 71.31 & 75.48 & 78.15 &  1 & 53.98 & 62.41 & 64.79 & 59.62 \\
        & Ours     & 75.97 & 78.62 & 85.11 &  0 & 59.66 & 57.57 & 59.85 & 54.68 \\
      \addlinespace[0.4ex]
      \multirow{2}{*}{QwQ-32B\textsuperscript{\ding{174}}}
        & Baseline & 70.33 & 75.51 & 75.42 &  17 & 55.05 & 66.37 & 68.47 & 62.37 \\
        & Ours     & 77.64 & 81.62 & 81.19 &  4 & 67.68 & 66.14 & 64.23 & 63.06 \\
      \addlinespace[0.4ex]
      \multirow{2}{*}{DeepSeek-R1-671B\textsuperscript{\ding{172}}}
        & Baseline & 69.40 & 73.13 & 78.01 &  0 & 49.74 & 60.13 & 62.69 & 56.96 \\
        & Ours     & 74.80 & 76.79 & \textbf{86.72} &  0 & 56.82 & 54.82 & 56.36 & 51.00 \\
      \addlinespace[0.4ex]
      \multirow{2}{*}{Qwen3-235B-A22B}
        & Baseline & 69.21 & 75.40 & 72.93 &  0 & 62.18 & 69.77 & 65.16 & 66.23 \\
        & Ours     & 78.16 & 81.53 & 83.40 &  0 & 66.09 & 63.06 & 62.27 & 60.18 \\
      \bottomrule
    \end{tabularx}%
  }
  \vspace{-5pt}
\end{table*}

\noindent\textbf{Evaluation Metrics.} \label{evaluation_metrics}
We adopt standard evaluation metrics to assess model performance, including accuracy, precision, recall, F1 score, the number of invalid outputs, and step-wise accuracy. Accuracy reflects the overall correctness, while precision and recall measure the proportion of correct predictions among all valid outputs and among all answerable questions, respectively. The F1 score provides a balanced summary of these two metrics. \textbf{Invalid outputs} refer to the number of samples from which no valid answer can be extracted using either instruction-based or rule-based decoding; final judgments on invalidity are made through semantic evaluation using GPT-4o. This design accommodates models specialized in specific tasks, aiming to explore whether they truly possess the ability to identify erroneous questions. Step-wise accuracy measures the correctness of each individual step executed in the MathQ-Verify framework. \textbf{S1\&2 to S5} represent the accuracy of predictions for Question correctness at each steps in the ValiMath. Notably, under comparable recall, our method consistently achieves higher precision, indicating stronger robustness in identifying valid questions and resisting spurious outputs.

\noindent\textbf{Baseline.}
For the analysis of invalid samples on the MathClean-Benchmark, we adhere to the evaluation methodology established in the original work. For the ValiMath dataset, we construct a baseline method that assesses the correctness of each input question directly, without employing the decomposition or multi-step verification procedures integral to the MathQ-Verify framework. This baseline provides a point of comparison to evaluate the efficacy of our proposed pipeline.

\noindent\textbf{Settings.}
All open-source models are deployed using the vLLM framework, while closed-source models are accessed via their respective official APIs. For open-source models, we adopt consistent settings: \texttt{temperature=0.7}, \texttt{top\_p=0.8}, \texttt{max\_tokens=16384}, \texttt{top\_k=20}. For math-specialized models (e.g., Qwen2.5-Math-7B), we set \texttt{max\_tokens=4096} to optimize performance. All experiments are conducted on 8 x NVIDIA H200 GPUs.

\vspace{-1em}
\subsection{Results and Analysis}
To address \textbf{Q1}, which investigates the effectiveness of MathQ-Verify, we conduct comprehensive evaluations on three benchmark datasets: MathClean-Simple, MathClean-Challenging, and ValiMath. The primary objective is to quantify the extent to which MathQ-Verify improves prediction accuracy and to assess its applicability across diverse model architectures. As presented in Table~\ref{tab:gsm8k_math}, our method yields substantial improvements in both accuracy and the reduction of invalid outputs across a wide range of models. In particular, models such as Qwen2.5-7B and LLaMA-3.1-8B exhibit notable gains in precision and F1 scores, which reflects the likelihood that predicted samples contain \textbf{truly correct data}. However, certain specialized models such as Qwen2.5-Math-7B demonstrate limitations in instruction-following capabilities, resulting in increased invalid outputs. These observations suggest that models with sufficient instruction-following capabilities generally exhibit performance improvements when augmented with MathQ-Verify.
\vspace{-0.1em}
\begin{table}[h]
\captionsetup{justification=centering}
\centering
\caption{Different model ensembles for $(n,k)$ configurations. \\
TP indicates correct positive predictions, while FP indicates incorrect positive predictions.}
\label{tab:best_ensembles}
\vspace{-1em}
\resizebox{0.48\textwidth}{!}{
\begin{tabular}{c c c c c c c}
    \toprule
    \textbf{$(n,k)$} & \textbf{Voters} & \textbf{Precision (\%)} & \textbf{Recall (\%)} & \textbf{F1 (\%)} & \textbf{TP} & \textbf{FP} \\
    \midrule
    (1, 1)   & \ding{172}             & 86.72 & 68.90 & 76.79 & 895  & 137 \\
    \midrule
    (2, 1)   & \ding{172} \ding{173} & 83.19 & 79.21 & 81.15 & 1029 & 208 \\
    (2, 2)   & \ding{172} \ding{173} & 89.56 & 62.74 & 73.79 & 815  & 95  \\
    \midrule
    (3, 1)   & \ding{172} \ding{173} \ding{174} & 78.42 & \textbf{86.99} & \textbf{82.48} & 1130 & 311 \\
    (3, 2)   & \ding{172} \ding{173} \ding{174} & 86.20 & 75.98 & 80.77 & 987  & 158 \\
    (3, 3)   & \ding{172} \ding{173} \ding{175} & 91.42 & 61.51 & 73.54 & 799  & 75  \\
    \midrule
    (4, 4)   & \ding{172} \ding{173} \ding{175} \ding{176} & 92.19 & 53.58 & 67.77 & 696  & 59  \\
    (5, 5)   & \ding{172} \ding{173} \ding{175} \ding{176} \ding{177} & \textbf{92.44} & 50.81 & 65.57 & 660  & 54  \\
    \bottomrule
\end{tabular}
}
\vspace{-0.7em}
\end{table}
To further evaluate the robustness and generalizability of MathQ-Verify, we extend our analysis beyond the MathClean benchmarks to the ValiMath dataset. This investigation aims to determine whether the observed performance gains persist under distributional shifts and dataset variations. As illustrated in Table~\ref{tab:eight_metrics_beautified}, MathQ-Verify consistently enhances performance across datasets. For example, GPT-4o and S1.1-32B achieve an improvement of 5.78 and 6.02 percentage points in F1 score, respectively, alongside increases in precision. These findings affirm the framework’s ability to generalize across diverse data distributions, effectively improving the reliability and precision of model outputs while mitigating spurious generations.

Furthermore, we conducted a stepwise evaluation from S1 to S5 to assess the contribution of each step to overall performance. For example, S3 means we directly evaluate the accuracy of the "Atomic Condition Error" on the entire dataset to assess the ability of models capturing this type of error, rather than evaluate a sample from S1 to S3 and calculate a multi-step accuracy. This evaluation aimed to understand how the various stages of the pipeline interact with different model and how they influence performance. The results indicate that both non-reasoning and reasoning models exhibit improvements in the early stages (S1\&2 and S3), with notable gains observed in models such as Qwen2.5-7B. However, reasoning models, such as QwQ-32B, showed a decline in accuracy during the later stages (S4 and S5). Note that the decrease here refers to performance relative to the baseline in detecting this error type; it shows that some models perform worse with more stepwise hints. S5 reflects only the accuracy of directly detecting this error, unrelated to full question verification accuracy. This step-wise analysis provides insights into how the every step in MathQ-Verify performs capturing different kinds of errors.

\begin{figure*}[t]
  \centering
  \includegraphics[width=0.9\linewidth]{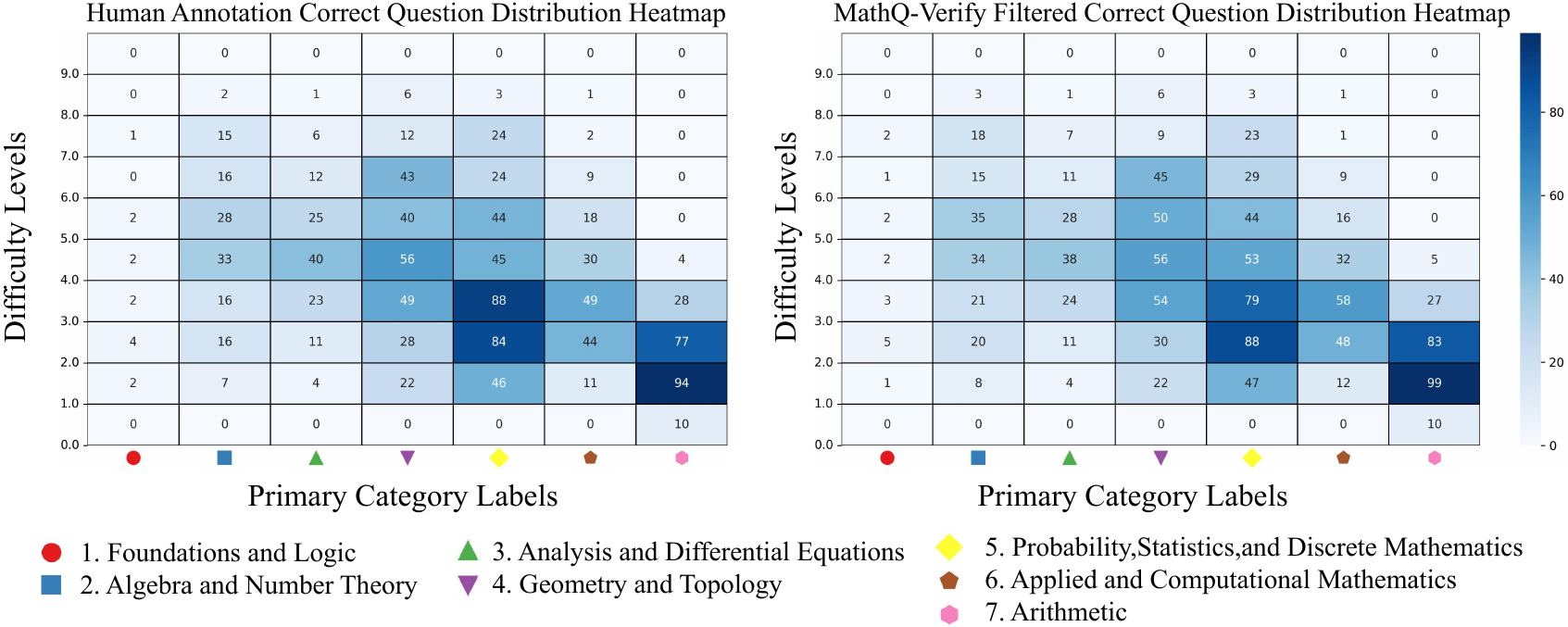}
  \vspace{-1em}
  \caption{Distribution heatmaps of correct questions across difficulty levels and primary mathematical categories. The left heatmap shows human-annotated correct question distribution, while the right heatmap shows the distribution of questions validated as correct by MathQ-Verify after filtering.}
  \label{fig:distribution}
  \vspace{-4mm}
\end{figure*}
\subsection{Multi-Model Voting Ensemble Trade-Offs.}
In this subsection we aim to address \textbf{Q2}, The numbers correspond to the models as indicated in Table~\ref{tab:eight_metrics_beautified}.

To prioritize \textbf{precision}, we observe that increasing the voting threshold ($k$) leads to more conservative predictions. For instance, in the $(3,3)$ configuration, where unanimous agreement is required, precision peaks at 91.42\%. However, this comes with a substantial drop in recall (61.51\%), indicating that many potentially valid samples are filtered out. 

Conversely, to maintain \textbf{recall}, configurations with lower $k$ values, such as $(3,1)$, offer a more balanced trade-off. This configuration achieves the highest F1 score (82.48\%) alongside a recall of 86.99\%, demonstrating its effectiveness in preserving a greater portion of correct samples while still enhancing prediction quality, as shown in Table~\ref{tab:best_ensembles}.

More voters means the increase of cost. Considering the \textbf{trade-offs between precision and recall}, and the objective of retaining a broad and diverse set of high-quality samples, the $(2,2)$ configuration emerges as the optimal choice. It provides an effective balance, enhancing question verification while ensuring that recall remains acceptable. Different configurations support varying precision and cost needs. This approach highlights the flexibility of $(n,k)$ voting, enabling us to prioritize precision without sacrificing recall, and ultimately improving the overall performance of the pipeline.

\vspace{-0.8em}
\subsection{Consistency of Data Distribution}
To answer \textbf{Q3}, which examines whether MathQ-Verify introduces significant distributional bias, we compare the joint category and difficulty distributions of the original human-annotated correct questions with those filtered by MathQ-Verify, as shown in Figure~\ref{fig:distribution}. The heatmaps reveal that the filtered distribution closely matches the original, with only minor shifts observed in certain categories and difficulty levels. Low-density areas, such as extreme difficulty levels and peripheral categories, remain consistent in both distributions. These results indicate that MathQ-Verify preserves the original category–difficulty structure, maintaining consistency with the human-annotated distribution.

\subsection{Ablation Study}
To evaluate each step in the MathQ-Verify pipeline, we perform an ablation study by removing one verification module at a time. Table~\ref{tab:ablation} reports results with GPT-o4-mini, where precision and F1 show the effect of each removal. Omitting the first two validation steps causes the largest drop—precision and F1 decrease by over 6\% and 3\% respectively. Removing contradiction detection slightly lowers performance by about 2\% in precision. Removing condition completeness increases F1 but slightly reduces precision, indicating a trade-off between strictness and coverage. Each module contributes uniquely, and their combination achieves the best precision–recall balance.

\vspace{-3pt}
\begin{table}[h]
  \centering
  \caption{Ablation study.}
  \vspace{-6pt}
  \label{tab:ablation}
  \resizebox{0.65\linewidth}{!}{%
    \begin{tabular}{@{}l r@{ }l r@{ }l@{}}
      \toprule
      \textbf{Variant} 
        & \multicolumn{2}{c}{\textbf{Precision (\%)}} 
        & \multicolumn{2}{c}{\textbf{F1 (\%)}} \\
      \midrule
      Baseline   
        & 78.86  &                 
        & 77.59  &               \\
      MathQ-Verify 
        & 80.88  & {\color{green!90}\(\uparrow2.02\)}  
        & 83.36  & {\color{green!90}\(\uparrow5.77\)}  \\
      \midrule
      w/o Step 1 
        & 78.41  & {\color{red!60}\(\downarrow2.47\)}  
        & 81.81  & {\color{red!30}\(\downarrow1.55\)} \\
      w/o Step 2 
        & 74.50  & {\color{red!90}\(\downarrow6.38\)}  
        & 80.00  & {\color{red!90}\(\downarrow3.36\)} \\
      w/o Step 3 
        & 80.19  & {\color{red!30}\(\downarrow0.69\)}  
        & 82.84  & {\color{red!30}\(\downarrow0.52\)} \\
      w/o Step 4 
        & 78.91  & {\color{red!60}\(\downarrow1.97\)}  
        & 83.13  & {\color{red!60}\(\downarrow0.23\)} \\
      w/o Step 5 
        & 79.53  & {\color{red!30}\(\downarrow1.35\)}  
        & 83.93  & {\color{green!60}\(\uparrow0.57\)} \\
      \bottomrule
    \end{tabular}%
  }
  \vspace{-4mm}
\end{table}

\section{Conclusion}

In this paper, we introduce a five steps math question-filtering pipeline in LLM training data, in contrast to prior work that predominantly filters based on answer correctness. To facilitate fine‐grained, step‐wise validation, we also construct the ValiMath benchmark, comprising 2,147 human‐verified examples. It is the first publicly released benchmark that performs step‑by‑step error checking on mathematical questions; every instance is annotated by multiple domain experts, providing new insights for the open‑source community on how to curate higher‑quality datasets. Experiments on different LLMs show that a single MathQ-Verify model achieves state‐of‐the‐art performance on different benchmarks. Both our five‑stage filtering pipeline and the expert‑annotated ValiMath benchmark offer the community a turnkey toolset, a scalable way to cleanse existing corpora and a gold‑standard testbed for validity‑aware model development.


\begin{acks}
    This work is supported by the National Natural Science Foundation of China (92470121, 62402016), National Key R\&D Program of China (2024YFA1014003), Humanities and Social Sciences Research Planning Fund Project of the Ministry of Education: "Research on Metacognitive Diagnosis Theory and Technology Driven by Multimodal Learning Data" (23YJA880091), and High-performance Computing Platform of Peking University.
\end{acks}

\bibliographystyle{ACM-Reference-Format}
\balance
\bibliography{sample-base}
\appendix

\section{Additional Implement Details} \label{additional_detail}
\subsection{Question Labels}

\vspace{0.5em}
We use GPT-4o to add the \textbf{following question labels}:

\vspace{0.5em}
\textbf{(1) Categorization:} We adopt the Mathematics Subject Classification (MSC) taxonomy~\citep{MSC2020}, which provides an extensive set of categories for mathematical topics. Given the complexity and large number of categories within MSC, we collaborated with domain experts to create a refined hierarchical classification scheme. The original categories were condensed into 7 primary categories, ensuring clear, non-overlapping coverage, and we introduced several secondary categories to capture more specific distinctions.

\begin{figure}[h]
\centering
\begin{minipage}{\columnwidth}
\lstset{
  basicstyle=\ttfamily\footnotesize,
  frame=single,
  columns=fullflexible,
  keepspaces=true
}
\begin{lstlisting}
1. Foundations and Logic
|-- 1.1 Mathematical Logic and Set Theory
|   (MSC 03, 06: propositional & predicate logic, set theory, 
ordered algebraic structures)
+-- 1.2 Basic Theory, Formalization, and History & Education
    (MSC 00, 01, part of 18, 97: foundational math, intro category
    theory, history, education)

2. Algebra and Number Theory
|-- 2.1 Linear Algebra and Group Theory            (MSC 15, 20)
|-- 2.2 Ring, Field, and Polynomial Algebra        (MSC 12, 16)
|-- 2.3 Commutative & Homological/Categorical      (MSC 13)
|-- 2.4 Number Theory                              (MSC 11)
+-- 2.5 Algebraic Geometry                         (MSC 14)

3. Analysis and Differential Equations
|-- 3.1 Real / Measure / Functional Analysis       (MSC 26, 28, 46)
|-- 3.2 Complex Analysis & Special Functions       (MSC 30, 33)
|-- 3.3 Differential Equations & Dynamical Sys.    (MSC 34, 35, 37)
|-- 3.4 Integral & Difference Equations            (MSC 39-45)
+-- 3.5 Harmonic Analysis                          (MSC 42, 43)

4. Geometry and Topology
|-- 4.1 Euclidean, Convex, Discrete Geometry       (MSC 51, 52)
|-- 4.2 Differential Geometry & Manifolds          (MSC 53, 57)
+-- 4.3 Topology & Algebraic Topology              (MSC 54, 55)

5. Probability, Statistics, and Discrete Math
|-- 5.1 Probability & Stochastic Processes         (MSC 60)
|-- 5.2 Mathematical Statistics                    (MSC 62)
+-- 5.3 Combinatorics & Graph Theory               (MSC 05)

6. Applied and Computational Mathematics
|-- 6.1 Numerical Analysis & Computation           (MSC 65)
|-- 6.2 Control, Variational, Optimization         (MSC 49)
|-- 6.3 OR & Game Theory                           (MSC 90, 91)
|-- 6.4 Systems Theory & Control                   (MSC 93)
|-- 6.5 Computer Science & Algorithms              (MSC 68)
|-- 6.6 Math Physics & Eng. Math                   (MSC 70, 74)
|-- 6.7 Information & Communication                (MSC 94)
+-- 6.8 Biomathematics                             (MSC 92)

7. Arithmetic
|-- 7.1 Basic Arithmetic and Number Operations
+-- 7.2 Word Problems and Real-Life Applications
\end{lstlisting}
\end{minipage}
\caption{Refined hierarchy of mathematical subjects aligned with MSC 2020 codes.}
\label{fig:msc-hierarchy}
\vspace{-0.5em}
\end{figure}

To guarantee both exhaustive coverage of contemporary mathematics and *minimal semantic
overlap* between labels, we treated the MSC codes as a controlled vocabulary and redesigned
them with tools borrowed from ontology engineering and faceted‑classification theory.

\begin{enumerate}[label=(\alph*)]
  \item \textbf{Concept consolidation.}  We first performed a formal concept analysis
        (FCA) over the MSC lattice, merging classes whose intension is identical but whose
        extension differs only in routine context (e.g., Banach– vs.\ Hilbert–space
        operator theory).  This step follows the
        ``least common subsumer'' principle of description‑logic ontologies.
  \item \textbf{Data‑driven similarity check.}  On a corpus of MathSciNet abstracts we
        computed distributional similarity between MSC codes (TF–IDF vectors with
        cosine distance).  Pairs whose linguistic contexts were highly similar were
        candidates for fusion, while strongly dissimilar pairs were forced into distinct
        secondary categories, thus reducing redundancy in the final tree.
  \item \textbf{Expert adjudication.}  Every automatic merge or split was independently
        reviewed by two subject experts; disagreements were resolved through a short
        Delphi round until consensus was reached.  This hybrid pipeline preserves
        domain fidelity while avoiding purely heuristic clustering.
  \item \textbf{Polyhierarchy minimisation.}  Except for a small, explicitly documented
        set of inherently cross‑cutting topics (e.g., homological algebra, which links
        algebra and topology), each secondary node is assigned a \emph{unique} parent.
        This “single inheritance unless necessary’’ rule mirrors the unique‑path
        restriction of the UMLS metathesaurus and greatly simplifies downstream
        indexing and retrieval.
  \item \textbf{Empirical coverage test.}  A pilot annotation of recent \texttt{arXiv}
        submissions showed that every paper mapped to at least one secondary category
        and that duplicate assignments were rare, indicating that the hierarchy is both
        comprehensive and well‑discriminated.
  \item \textbf{Scope notes and maintenance.}  For each secondary category we wrote a
        short scope note that states inclusion and exclusion criteria and lists the
        original MSC ranges it subsumes.  These notes will be version‑controlled to
        support future revisions as mathematical research evolves.
\end{enumerate}

Combining algorithmic consolidation with expert validation thus yields a lean yet
expressive hierarchy that remains faithful to the MSC while offering clearer semantics
for large‑scale annotation and analysis.

\textbf{(2) Difficulty Classification:} To classify the difficulty level of mathematical questions, we use the method of OMNI-MATH~\citep{gao2024omnimathuniversalolympiadlevel}. Since shot-level supervision was insufficient, we manually curated specific representative exemplars to ensure accurate delineation of difficulty levels for the dataset, as shown in Figure~\ref{fig:difficulty-scale}. This manual curation enabled us to better define the boundaries between different difficulty levels, facilitating more reliable analysis.

\subsection{Handling the Invalid response from model}

Although we explicitly specify the required output format during generation, models of varying sizes, capabilities, and contextual understanding do not always conform to the expected structure. In practice, most models produce only a small number of format violations, which can still be reliably handled via regular expressions to extract the final answers.

However, for math-specialized models, it is normal to observe a significantly higher incidence of unstructured or malformed outputs. These outputs often deviate from the expected format in ways that are not easily recoverable through rule-based methods, making it difficult to extract the final answer even with robust regular expression patterns.

To address this, we incorporate GPT-4o as a semantic parser to interpret and resolve ambiguous or non-compliant outputs. Specifically, for any model response that fails standard parsing, GPT-4o is prompted to infer the intended final answer based on the full output context. This ensures that downstream evaluation remains consistent and robust across diverse model types.

\begin{figure}[h]
\centering
\begin{minipage}{\columnwidth}
\lstset{
  basicstyle=\ttfamily\footnotesize,
  frame=single,
  columns=fullflexible,
  keepspaces=true,
  breaklines=true          
}
\begin{lstlisting}[language=]
Difficulty-level overview
  1.0 : Lower elementary - basic arithmetic and simple counting
  1.5 : Harder elementary contest - digit properties, easy divisibility
  2.0 : Easy probability / combinatorics - single-step numeric answer
  2.5 : Moderate algebra / mean problems - multi-step manipulation
  3.0 : Basic geometry constructions & area - standard secondary facts
  3.5 : Early AIME tier - logarithms or equations with light reasoning
  4.0 : Upper-middle contest - recurrences or piecewise cases
  4.5 : USAJMO proof - perfect-square arguments in number theory
  5.0 : JBMO tier - symmetry, systems of equations, higher algebra
  5.5 : Final AMC12 problems - trigonometry, triangle centers, extrema
  6.0 : Upper AIME - combined geometry and algebra, detailed computation
  6.5 : USAMO proof - concurrency or geometric transformations
  7.0 : Intro IMO proof - set constructions, algebra-geometry blend
  7.5 : Advanced USAMO - functional equations, heavy algebra
  8.0 : Tough IMO problems - limit partitions or hard combinatorial proofs
  8.5 : Complex IMO geometry - circle chains, high tangency
  9.0 : Hard IMO combinatorics / number theory - uniqueness, symmetry
  9.5 : Top IMO constructions - large arrays, intricate configurations
 10.0 : Ultimate IMO finale - deep limits and algebraic geometry
\end{lstlisting}
\end{minipage}
\caption{Difficulty scale adopted for dataset annotation.}
\label{fig:difficulty-scale}
\end{figure}

\section{Annotation Guidelines} \label{appendix:annotation}

The following guidelines were provided to expert annotators to ensure consistent, accurate, and interpretable annotations throughout the MathQ-Verify validation process. Each question in the dataset was assessed under a structured five-stage framework, and annotators were trained to identify the earliest step at which an issue arises. For all questions deemed incorrect, annotators were also encouraged to include clarifying remarks.

\paragraph{Task Objective.}
For each question $q$, annotators determined whether it is fully valid and solvable. A valid question must (i) be mathematically framed, (ii) be linguistically well-formed, (iii) consist of correct atomic premises, (iv) be globally consistent, and (v) be complete with respect to its goals. If a failure occurred at any of these stages, the question was marked \texttt{incorrect}, and the earliest failing step was recorded. This step-level diagnostic approach enables fine-grained error tracking and helps inform both dataset filtering and model evaluation.

\paragraph{Stage-by-Stage Evaluation.}
Each question was assessed in the following order. Below we describe the core criteria and provide examples of typical failures encountered.

\begin{itemize}
    \item \textbf{Stage 1 — Instruction Validity:} A valid question should contain a clear mathematical prompt with no meta instructional language (e.g., “rewrite this” or “solve this for me”) and no embedded answers. For example,  
    \textit{Invalid:} “Please rewrite this problem so it sounds more natural.”  
    \textit{Invalid:} “The answer is $x=3$. How did we get this?”  
    \textit{Valid:} “Find all real numbers $x$ such that $2x + 5 = 11$.”  
    This step filters out synthetic or contaminated prompts that would bias a solver or are not true standalone math questions.

    \item \textbf{Stage 2 — Linguistic Cleanliness:} The question must be syntactically and semantically well-formed. For instance, questions like  
    \textit{Invalid:} “Let $x$ be an integer such that $x > 2$.” (missing closing dollar sign)  
    \textit{Invalid:} “If $a$ is prime and a + 1 is evn, prove it.” (typo: “evn”)  
    While surface typos are tolerated when they don't impair understanding, malformed symbols or broken expressions (especially in LaTeX) often render the question unusable and are flagged at this stage.

    \item \textbf{Stage 3 — Atomic Validity:} Here, annotators inspect each atomic premise in the logical decomposition $\mathcal{P}(q)$ and verify its correctness.  
    \textit{Invalid:} “Let $x$ be both an integer and a real number between 2.5 and 2.8.” (contradiction in type)  
    \textit{Invalid:} “Define $A$ as the matrix with elements $a_{ij} = i/j$ for $i=1$ to $n$, $j=0$ to $n$.” (division by zero)  
    \textit{Valid:} “Let $f(x) = \sqrt{x^2 + 1}$ and define $g(x) = f(f(x))$.”  
    Even if the overall question seems plausible, a single broken assumption here invalidates the entire construction.

    \item \textbf{Stage 4 — Consistency:} The union of all premises must be logically coherent. For example:  
    \textit{Invalid:} “Let $x$ be a prime number greater than 10 and divisible by 4.”  
    Each statement in isolation may seem correct, but their combination creates a contradiction. Annotators often need to reason jointly across premises here—e.g., verify that constraints on ranges, algebraic structure, and definitions don't conflict.

    \item \textbf{Stage 5 — Completeness:} Annotators test whether the question provides enough information to reach the goal.  
    \textit{Invalid:} “In triangle $ABC$, angle $A=60^\circ$, and $AB=5$. Find the area.” (incomplete: more side/angle info needed)  
    \textit{Valid:} “In triangle $ABC$, angle $A=60^\circ$, $AB=5$, and $AC=5$. Find the area.” (now solvable via standard formula)  
    Annotators are instructed to simulate solving the question: if essential steps are blocked due to missing facts, the item is labeled \texttt{incorrect} at this stage.
\end{itemize}

\paragraph{Annotation Format.}
For each question, the final annotation includes:
\begin{itemize}
    \item A binary correctness label: \texttt{correct} or \texttt{incorrect}.
    \item If incorrect: the name of the earliest failed stage (e.g., “Atomic Validity”).
    \item Optional remarks to explain the failure or clarify ambiguity.
\end{itemize}
These annotations were collected using a structured interface that forced annotators to proceed sequentially, ensuring consistency in stage ordering and reducing oversights.

\paragraph{Consensus and Conflict Resolution.}
Each question was independently labeled by at least two experts. If labels disagreed, a reconciliation meeting was held to discuss the discrepancy and reach a consensus. Annotators were encouraged to refer back to the guidelines, prior examples, and formal definitions to justify their decision. For persistent disagreements, a senior adjudicator provided the final call after consulting both parties.

\paragraph{Inter-Annotator Reliability.}
To ensure labeling stability, we conducted inter-annotator agreement studies across random subsets of the data. Annotators demonstrated high consistency across clean and noisy examples, with confusion concentrated near ambiguous boundaries (e.g., whether a linguistic error meaningfully distorts semantics). These reliability metrics, along with per-stage label statistics, are reported in the main paper.

\paragraph{Ethical and Procedural Standards.}
All annotators underwent a month-long training program and were evaluated not only for mathematical knowledge but also for consistency, fairness, and ethical integrity. They were instructed to avoid over-judgment, abstain from guessing intent, and follow formal criteria even when the question “seems understandable.” Ambiguous or borderline cases were flagged for collective discussion. This ensured that all annotations reflect reasoned, transparent, and replicable decisions.


\section{The Algorithm of MathQ-Verify}\label{appendix:algo}

\vspace{-1em}
\begin{algorithm}[h]
\caption{MathQ-Verify}\label{alg:validate}
\KwIn{ Question $q \in \mathcal{Q}$}
\KwOut{$V(q)\in\{0,1\}$}

$\InstValid \gets
  \IsMathQ(q)\land\lnot\HasMisleadingCue(q)\land\lnot\HasAnswerLeak(q)$\;
\quad \quad \lIf{$\lnot\InstValid$}{\Return 0}

$\Clean \gets
  \lnot\LinguisticError(q)\land\exists\,(\mathcal P(q),\mathcal G(q))$\;
\quad \quad \lIf{$\lnot\Clean$}{\Return 0}

$\AtomValidAll \gets
  \forall P_j\!\in\!\mathcal P(q):\ \AtomValid(P_j)$\;
\quad \quad \lIf{$\lnot\AtomValidAll$}{\Return 0}

$\Consistent \gets \Sat\!\bigl(\mathcal P(q)\bigr)$\;
\quad \quad \lIf{$\lnot\Consistent$}{\Return 0}

$\Complete \gets
  \forall G\!\in\!\mathcal G(q):\ \mathcal P(q)\models G$\;
\quad \quad \lIf{$\lnot\Complete$}{\Return 0}

\quad \Return 1\tcp*{all checks passed}
\end{algorithm}

\end{document}